\documentclass[runningheads]{llncs}
\usepackage{graphicx}
\usepackage{graphicx}
\usepackage{amsmath,amssymb} 
\usepackage{color}
\usepackage{booktabs}
\usepackage{multirow}
\usepackage{amssymb}
\usepackage{bbm}
\usepackage{algorithmicx,algorithm}
\usepackage[noend]{algpseudocode}
\usepackage{mathrsfs}
\newcommand{\ie}{\textit{i}.\textit{e}., }
\newcommand{\eg}{\textit{e}.\textit{g}., }
\newcommand{\etc}{\textit{e}\textit{t}\textit{c}}
\newcommand{\myPara}[1]{\vspace{0.02in}\noindent\textbf{#1}}
\usepackage{pifont}
\begin{document}
\title{HiCo: Hierarchical Contrastive Learning for Ultrasound Video Model Pretraining}
\titlerunning{HiCo: Hierarchical Contrastive Learning}

\author{Chunhui Zhang\inst{1,2}\orcidID{0000-0002-9017-1828} \and
Yixiong Chen\inst{3,4}\orcidID{0000-0003-0268-076X} \and
Li Liu\inst{3,4,}\thanks{Corresponding author.}\orcidID{0000-0002-4497-0135} \and
Qiong Liu\inst{2}\orcidID{0000-0002-5808-2761} \and
Xi Zhou\inst{2,1}\orcidID{0000-0001-9943-5482}}
\authorrunning{Chunhui Zhang et al.}
\institute{
Shanghai Jiaotong University, 200240 Shanghai, China \and
CloudWalk Technology Co., Ltd, 201203 Shanghai, China \and
The Chinese University of Hong Kong (Shenzhen), 518172 Shenzhen, China
\email{liuli@cuhk.edu.cn} \and
Shenzhen Research Institute of Big
Data, 518172 Shenzhen, China
}
\maketitle              % typeset the header of the contribution
\begin{abstract}
The self-supervised ultrasound (US) video model pretraining can use a small amount of labeled data to achieve one of the most promising results on US diagnosis. However, it does not take full advantage of multi-level knowledge for learning deep neural networks (DNNs), and thus is difficult to learn transferable feature representations. This work proposes a hierarchical contrastive learning (HiCo) method to improve the transferability for the US video model pretraining. HiCo introduces both peer-level semantic alignment and cross-level semantic alignment to facilitate the interaction between different semantic levels, which can effectively accelerate the convergence speed, leading to better generalization and adaptation of the learned model. Additionally, a softened objective function is implemented by smoothing the hard labels, which can alleviate the negative effect caused by local similarities of images between different classes. Experiments with HiCo on five datasets demonstrate its favorable results over state-of-the-art approaches. The source code of this work is publicly available at \url{https://github.com/983632847/HiCo}.

%\keywords{First keyword  \and Second keyword \and Another keyword.}
\end{abstract}
\section{Introduction}
Thanks to the cost-effectiveness, safety, and portability, combined with a reasonable sensitivity to a wide variety of pathologies, ultrasound (US) has become one of the most common medical imaging techniques in clinical diagnosis~\cite{born2021accelerating}. To mitigate sonographers' reading burden and improve diagnosis efficiency, automatic US analysis using deep learning is becoming popular~\cite{gao2016describing,chen2021uscl,liu2020semi,gao2021multi}. In the past decades, a successful practice is to train a deep neural network (DNN) on a large number of well-labeled US images within the supervised learning paradigm~\cite{born2021accelerating,su2020convolutional}. However, annotations of US images and videos can be expensive to obtain and sometimes infeasible to access because of the expertise requirements and time-consuming reading, which motivates the development of US diagnosis that requires few or even no manual annotations.

\begin{figure}[t]
\centering
\includegraphics[width=120mm]{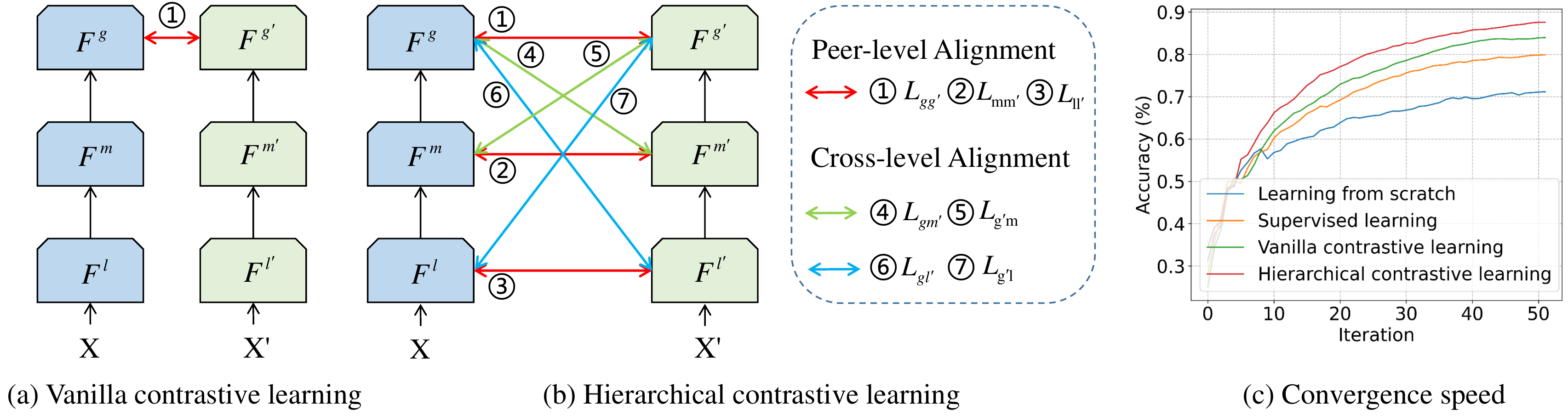} 
\caption{Motivation of hierarchical contrastive learning. Unlike (a) vanilla contrastive learning, our (b) hierarchical contrastive learning can fully take advantage of both peer-level and cross-level information. Thus, (c) the pretraining model from our proposed hierarchical contrastive learning can accelerate the convergence speed, which is much better than learning from scratch, supervised learning, and vanilla contrastive learning.} 
\label{fig:Motivation}
\end{figure}

In recent years, pretraining combined with fine-tuning has attracted great attention because it can transfer knowledge learned on large amounts of unlabeled or weakly labeled data to downstream tasks, especially when the amount of labeled data is limited. This has also profoundly affected the field of US diagnosis, which started to pretrain models from massive unlabeled US data according to a pretext task. To learn meaningful and strong representations, the US video pretraining methods are designed to correct the order of a reshuffled video clip, predict the geometric transformation applied to the video clip or colorize a grayscale image to its color version equivalent~\cite{jiao2020self,zhang2016colorful}. Inspired by the powerful ability of contrastive learning (CL)~\cite{he2020momentum,chen2020simple} in computer vision, some recent studies propose to learn US video representations by CL~\cite{jiao2020self2,chen2021uscl}, and showed a powerful learning capability~\cite{zhang2020contrastive,jiao2020self2}. However, most of the existing US video pretraining methods following the vanilla contrastive learning setting~\cite{chen2020simple,chen2020improved}, only use the output of a certain layer of a DNN for contrast (see Fig.~\ref{fig:Motivation}(a)). Although the CL methods are usually better than learning from scratch and supervised learning, the lack of multi-level information interaction will inevitably degrade the transferability of pretrained models~\cite{chen2021uscl,gao2020accurate}.

To address the above issue, we first propose a hierarchical contrastive learning (HiCo) method for US video model pretraining. The main motivation is to design a \emph{feature-based} peer-level and cross-level semantic alignment method (see Fig.~\ref{fig:Motivation}(b)) to improve the efficiency of learning and enhance the ability of feature representation. Specially, based on the assumption that the top layer of a DNN has strong semantic information, and the bottom layer has high-resolution local information (\eg texture and shape)~\cite{ren2015faster}, we design a joint learning task to force the model to learn multi-level semantic representations during the CL process: minimize the peer-level semantic alignment loss (\ie \textcircled{1} global CL loss, \textcircled{2} medium CL loss, and \textcircled{3} local CL loss) and cross-level semantic alignment loss (\ie \textcircled{4}, \textcircled{5} global-medium CL losses, and \textcircled{6}, \textcircled{7} global-local CL losses) simultaneously. Intuitively, our framework can greatly improve the convergence speed of the model (\ie providing a good initialized model for downstream tasks)~(see Fig.~\ref{fig:Motivation}(c)), due to the sufficient interaction of peer-level and cross-level information. Different from existing methods~\cite{xu2022seed,lee2020contrastive,li2022keywords,li2022hiclre,wang2022heloc}, this work assumes that the knowledge inside the backbone is sufficient but underutilized, so that simple yet effective peer-level and cross-level semantic alignments can be used to enhance feature representation other than designing a complex structure. In addition, medical images from different classes/lesions may have significant local similarities (\eg normal and infected individuals have similar regions of tissues and organs unrelated to disease), which is more severe than natural images. Thus, we follow the popular label smoothing strategy to design a \emph{batch-based} softened objective function during the pretraining to avoid the model being over-confident, which alleviates the negative effect caused by local similarities.

The main contributions of this work can be summarized as follows: 

1) We propose a novel hierarchical contrastive learning method for US video model pretraining, which can make full use of the multi-level knowledge inside a DNN via peer-level semantic alignment and cross-level semantic alignment. 

2) We soften one-hot labels during the pretraining process to avoid the model being over-confident, alleviating the negative effect caused by local similarities of images between different classes.

3) Experiments on five downstream tasks demonstrate the effectiveness of our approach in learning transferable representations.

%===========================================================
\section{Related Work}
\label{sec:relatedwork}
We first review related works on supervised learning for US diagnosis and then discuss the self-supervised representation learning.

\subsection{US Diagnosis}
With the rise of deep learning in computer vision, supervised learning became the most common strategy in US diagnosis with DNN~\cite{born2021accelerating,chen2021uscl,schmarje2020survey,he2016deep,gao2022efficient}. In the last decades, numerous datasets and methods have been introduced for US image classification~\cite{chi2017thyroid}, detection~\cite{yap2017automated} and segmentation~\cite{huang2017breast} tasks. For example, some US image datasets with labeled data were designed for breast cancer classification~\cite{Walid2019DB,Rodrigues2018Mendeley}, breast US lesions detection~\cite{yap2017automated}, diagnosis of malignant thyroid nodule~\cite{pedraza2015open,nguyen2020ultrasound}, and automated measurement of the fetal head circumference~\cite{li2020automated}. At the same time, many deep learning approaches have been done on lung US~\cite{kalafat2020lung,long2017lung}, B-line detection or quantification~\cite{kerdegari2021automatic,wang2019quantifying}, pleural line extraction~\cite{carrer2020automatic}, and subpleural pulmonary lesions~\cite{xu2020boundary}. Compared with image-based datasets, recent video-based US datasets~\cite{born2021accelerating,chen2021uscl} are becoming much richer and can provide more diverse categories and data modalities (\eg  convex and linear probe US images~\cite{chen2021uscl}). Thus, many works are focused on video-based US diagnosis within the supervised learning paradigm. In~\cite{born2021accelerating}, a frame-based model was proposed to correctly distinguish COVID-19 lung US videos from healthy and bacterial pneumonia data. Other works focus on quality assessment for medical US video compressing~\cite{6823625},  localizing target structures~\cite{kwitt2013localizing}, or describing US video content~\cite{gao2016describing}. Until recently, many advanced DNNs (\eg UNet~\cite{ronneberger2015u}, DeepLab~\cite{chen2014semantic,chen2018encoder}, Transformer~\cite{cao2021swin}), and technologies (\eg neural architecture search~\cite{weng2019unet}, reinforcement learning~\cite{huang2022extracting}, meta-learning~\cite{gong2021diagnosis}) have brought great advances in supervised learning for US diagnosis. Unfortunately, US diagnosis using supervised learning highly relies on large-scale labeled, often expensive medical datasets.

\subsection{Self-supervised Learning}

Recently, many self-supervised learning methods for visual feature representation learning have been developed without using any human-annotated labels~\cite{korbar2018cooperative,oord2018representation,ye2019unsupervised}. Existing self-supervised learning methods can be divided into two main categories, \ie learning via pretext tasks and CL. A wide range of pretext tasks have been proposed to facilitate the development of self-supervised learning. Examples include solving jigsaw puzzles~\cite{noroozi2016unsupervised}, colorization~\cite{zhang2016colorful}, image context restoration~\cite{chen2019self}, and relative patch prediction~\cite{doersch2015unsupervised}. However, many of these tasks rely on ad-hoc heuristics that could limit the generalization and robustness of learned feature representations for downstream tasks~\cite{chen2020improved,chen2020simple}. The CL has emerged as the front-runner for self-supervision representation learning and has demonstrated remarkable performance on downstream tasks. Unlike learning via pretext tasks, CL is a discriminative approach that aims at grouping similar positive samples closer and repelling negative samples. To achieve this, a similarity metric is used to measure how close two feature embeddings are. For computer vision tasks, a standard loss function, \ie Noise-Contrastive Estimation loss (InfoNCE)~\cite{gutmann2010noise}, is evaluated based on the feature representations of images extracted from a backbone network (\eg ResNet~\cite{he2016deep}). Most successful CL approaches are focused on studying effective contrastive loss, generation of positive and negative pairs, and sampling method~\cite{chen2020simple,he2020momentum}. SimCLR~\cite{chen2020simple} is a simple framework for CL of visual representations with strong data augmentations and a large training batch size. MoCo~\cite{he2020momentum}  builds a dynamic dictionary with a queue and a moving-averaged encoder. Other works explores learning without negative samples~\cite{grill2020bootstrap,chen2021exploring}, and incorporating self-supervised learning with visual transformers~\cite{chen2021empirical}, \etc.

Considering the superior performance of contrastive self-supervised learning in computer vision and medical imaging tasks, this work follows the line of CL. First, we propose both peer-level and cross-level alignments to speed up the convergence of the model learning, compared with the existing CL methods, which usually use the output of a certain layer of the network for contrast (see Fig.~\ref{fig:Motivation}). Second, we design a softened objective function to facilitate the CL by addressing the negative effect of local similarities between different classes.

%===========================================================
\section{Hierarchical Contrastive Learning}
\label{sec:Method}

In this section, we present our HiCo approach for US video model pretraining. To this end, we first introduce the preliminary of CL, after that present the peer-level semantic alignment and cross-level semantic alignment, and then describe the softened objective function. The framework of HiCo is illustrated in Fig.~\ref{fig:HiCo}.

\subsection{Preliminary} 
The vanilla contrastive learning learns a global feature encoder $\Phi$ and a projection head $g$ that map the image $X$ into a feature vector $\textbf{F}\!=\!g(\Phi(X))\in\mathbb{R}_{d}$ by minimizing the InfoNCE loss~\cite{gutmann2010noise}:
\begin{equation}
\mathscr L_{nce}{(\textbf{F}_i,\textbf{F}_j)} = -\log\frac{exp(sim(\textbf{F}_i, \textbf{F}_j)/\tau)}{\sum_{k=1}^{2N} \mathbbm{1}_{[k\neq i]}exp(sim(\textbf{F}_i,\textbf{F}_k)/\tau)}.  
\label{eq:vanilla_CL}
\end{equation}
where $(\textbf{F}_i, \textbf{F}_j)$ are the global feature vectors of the two views of image $X$, $N$ is the batch size. $sim(\textbf{F}_i, \textbf{F}_j)=\textbf{F}_i\cdot \textbf{F}_j/(||\textbf{F}_i||||\textbf{F}_j||)$ denotes the cosine similarity, $\mathbbm{1}_{k\neq i}\in\{0,1\}$ is an indicator function evaluating to 1 iff $k\neq i$, and $\tau$ is a tuning temperature parameter.

In practice, $\Phi$ is a DNN (\eg ResNet~\cite{he2016deep}) and the objective function is minimized using stochastic gradient descent. The two feature vectors of the one image are a positive pair, while the other $2(N-1)$ examples within a mini-batch are treated as negative examples. The positive and negative contrastive feature pairs are usually the global representations (\eg the output from the last layer of a DNN) that lack multi-scale information from different feature layers, which are important for downstream tasks. In addition, having a large number of negative samples is critical while minimizing the InfoNCE loss. Hence, a large batch size (\eg 4096 in SimCLR~\cite{chen2020simple}) is required during the pretraining.

\begin{figure}[t]
\centering
\includegraphics[width=120mm]{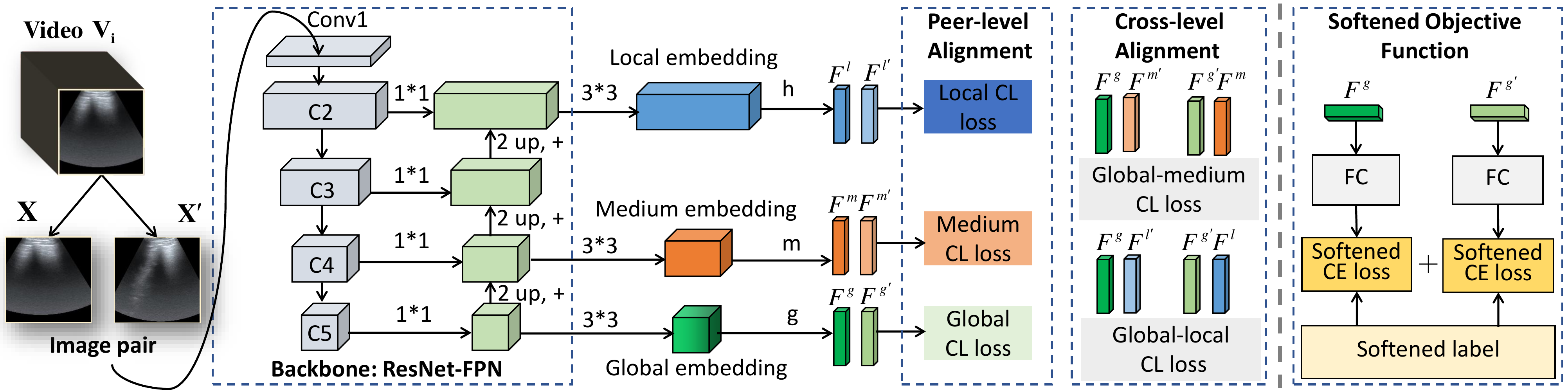} 
\caption{Overall framework of the proposed HiCo, which consists of peer-level semantic alignment, cross-level semantic alignment, and softened objective function. 1) We extract two images from each US video as a positive sample pair. 2) We use ResNet-FPN as the backbone to obtain the local, medium and global embeddings, followed by three projection heads $h$, $m$ and $g$. 3) The entire network is optimized by minimizing peer-level semantic alignment loss (\ie local CL loss, medium CL loss and global CL loss), cross-level semantic alignment loss (\ie global-medium CL loss and global-local CL loss), and the softened CE loss.}
\label{fig:HiCo}
\end{figure}

To address the above problem, we explore multi-level semantic alignments for CL, peer-level and cross-level ones (see Fig.\ref{fig:HiCo}). Specifically, we use ResNet-FPN as the backbone. For each image pair, we extract two local feature vectors ($\textbf{F}^{l}$ and $\textbf{F}^{l'}$), two medium feature vectors ($\textbf{F}^{m}$ and $\textbf{F}^{m'}$) and two global feature vectors ($\textbf{F}^{g}$ and $\textbf{F}^{g'}$) from Conv2 (C2), Conv4 (C4) and Conv5 (C5), respectively. We then optimize the peer-level and cross-level alignments simultaneously.

\subsection{Peer-level Semantic Alignment}

\subsubsection{Fine-grained Contrast.}

The $\textbf{F}^{l}$ and $\textbf{F}^{l'}$ encode the fine-grained local information (\eg edges and shapes) of original images. Such fine-grained information is useful for US diagnosis, but is usually ignored in existing CL algorithms. To leverage the fine-grained information, we define the local CL loss $\mathcal L^{local}_{ll'}$ as 
\begin{equation}
\mathcal L^{local}_{ll'} = \frac{1}{2N}\sum_{i=1}^{N}(\mathscr L_{nce}(\textbf{F}_{i}^{l}, \textbf{F}_{i}^{l'})+\mathscr L_{nce}(\textbf{F}_{i}^{l'}, \textbf{F}_{i}^{l})),  
\label{eq:local_CL}
\end{equation}
where $N$ denotes the batch size, $\mathscr L_{nce}(\cdot)$ is the InfoNCE loss~\cite{gutmann2010noise}.

\subsubsection{Medium-grained Contrast.}
Considering that $\textbf{F}^{m}$ and $\textbf{F}^{m'}$ capture medium-grained information of original images, we therefore define the medium CL loss $\mathcal L^{medium}_{mm'}$ as 
\begin{equation}
\mathcal L^{medium}_{mm'} = \frac{1}{2N}\sum_{i=1}^{N}(\mathscr L_{nce}(\textbf{F}_{i}^{m}, \textbf{F}_{i}^{m'})+\mathscr L_{nce}(\textbf{F}_{i}^{m'}, \textbf{F}_{i}^{m})),
\label{eq:medium_CL}
\end{equation}

Notably, we find that medium-grained information demonstrate complementary superiority relative to fine-grained and global information, further improving model performance (see Section \ref{sec:ablation} Table~\ref{table:Ablation_Study_peer_level}). 

\subsubsection{Coarse-grained Contrast.}
Owing to $\textbf{F}^{g}$ and $\textbf{F}^{g'}$ capture coarse-grained global information of original images, we hope to reach a consensus among their representations by maximizing the similarity between global embeddings from the same video, while minimizing the similarity between the global embeddings from different videos. Thus, the global CL loss $\mathcal L^{global}_{gg'}$ can be defined as 
\begin{equation}
\mathcal L^{global}_{gg'} = \frac{1}{2N}\sum_{i=1}^{N}(\mathscr L_{nce}(\textbf{F}_{i}^{g}, \textbf{F}_{i}^{g'})+\mathscr L_{nce}(\textbf{F}_{i}^{g'}, \textbf{F}_{i}^{g})), 
\label{eq:global_CL}
\end{equation}

\subsection{Cross-level Semantic Alignment}

\subsubsection{Global-local Contrast.} We regard the global feature vector $\textbf{F}^{g}$ as the \emph{anchor} of the local feature vector $\textbf{F}^{l'}$, because it contains the global semantic information of the original image and shares some semantic content with the local feature vector. Thus, we define the global-local objective $\mathcal L_{gl'}$ to make the local feature vectors move closer to the global ones as
\begin{equation}
\mathcal L_{gl'} = \frac{1}{2N}\sum_{i=1}^{N}(\mathscr L_{nce}(\textbf{F}_{i}^{g}, \textbf{F}_{i}^{l'})+\mathscr L_{nce}(\textbf{F}_{i}^{l'}, \textbf{F}_{i}^{g})),
\label{eq:global_local_CL}
\end{equation}

The $\mathcal L_{g'l}$ can be calculated similarly. Then, the global-local CL loss can be written as $\mathcal L^{global}_{local} =\mathcal L_{gl'}+\mathcal L_{g'l}$.

\setlength{\textfloatsep}{20pt}
\begin{algorithm}[t]
	\caption{Hierarchical Contrastive Learning} %算法的名字
	\hspace*{0.02in} {\bf Input:} %算法的输入， \hspace*{0.02in}用来控制位置，同时利用 \\ 进行换行
	US videos $\textbf{V}$, backbone $\Phi$, projection heads $g, m, h$, linear classifier $f_{\theta}$, hyper-parameters $\lambda, \alpha, \beta$, max epoch $e_{max}$, batch size $N$.\\
	\hspace*{0.02in} {\bf Output:} %算法的结果输出
	pretrained backbone $\Phi$.
	\begin{algorithmic}[1]
		\State random initialize $\Phi$, $g, m, h$, and $f_{\theta}$ % \State 后写一般语句
		\For{$e=1,2,...,e_{max}$}
		\For{random sampled US videos $\{\textbf{V}_i\}_{i=1}^{N}$} % For 语句，需要和EndFor对应
		\State \# extract two images from each US
video as a positive sample pair.
		\State random sample $\{(\textbf{x}_i^{(1)},\textbf{x}_i^{(2)}), \textbf{y}_i\}^N_{i=1}$ into mini-batch
		\For{$i \in \{1,...,N\}$} % For 语句，需要和EndFor对应
		\State \# Augment image pair for each US video.
		\State random cropping, resizing, ﬂipping, and color jitter
		
		\State \# Get local, medium and global embeddings.
		\State $\textbf{F}_{i}^{l}=h(\Phi(\textbf{x}_{i}^{(1)}))$, $\textbf{F}_{i}^{l'}=h(\Phi(\textbf{x}_{i}^{(2)}))$;
		
		\State $\textbf{F}_{i}^{m}=m(\Phi(\textbf{x}_{i}^{(1)}))$, $\textbf{F}_{i}^{m'}=m(\Phi(\textbf{x}_{i}^{(2)}))$;
		
		\State $\textbf{F}_{i}^{g}=g(\Phi(\textbf{x}_{i}^{(1)}))$, $\textbf{F}_{i}^{g'}=g(\Phi(\textbf{x}_{i}^{(2)}))$;
		
		\State  \# Get outputs of the linear classifier.
		\State $\textbf{o}_{i}= f_{\theta}(\textbf{F}_{i}^{g})$, $\textbf{o}_{i}'= f_{\theta}(\textbf{F}_{i}^{g'})$
		\EndFor
		\EndFor
		
		\State \# peer-level semantic alignment
		\For{$i \in \{1,...,N\}$}
		\State calculate the local CL loss $\mathcal L^{local}_{ll'}$ by Eq.~(\ref{eq:local_CL})

		\State calculate the medium CL loss $\mathcal L^{medium}_{mm'}$ by Eq.~(\ref{eq:medium_CL})
		
		\State calculate the global CL loss $\mathcal L^{global}_{gg'}$ by Eq.~(\ref{eq:global_CL})
		\EndFor

		\State \# cross-level semantic alignment
		\For{$i \in \{1,...,N\}$}
		\State calculate the global-local CL loss $\mathcal L^{global}_{local}$ by Eq.~(\ref{eq:global_local_CL})
		
		\State calculate the global-medium CL loss $\mathcal L^{global}_{medium}$
		\EndFor
		
		\State calculate the overall CL loss $\mathcal L^{con}$ by Eq.~(\ref{eq:overall_CL})

		\State calculate the softened CE loss $\mathcal L^{soften}$ by Eq.~(\ref{eq:classification})

		\State $\mathcal L = \mathcal L^{con} + \beta \mathcal L^{soften}$
		\State update $\Phi$, $g, m, h$, and $f_{\theta}$  through gradient descent
		\EndFor
		\State \Return pretrained $\Phi$, discard $g, m, h$, and $f_{\theta}$
	\end{algorithmic}
	\label{alg:algo}
\end{algorithm}

\subsubsection{Global-medium Contrast.} Similar to the global-local CL loss, the global-medium CL loss can be written as $\mathcal L^{global}_{medium} =\mathcal L_{gm'}+\mathcal L_{g'm}$.

Therefore, the overall CL loss of HiCo is formulated as
\begin{equation}
\mathcal L^{con} = \lambda (\mathcal L^{local}_{ll'} + \mathcal L^{medium}_{mm'} +\mathcal L^{global}_{gg'}) + (1-\lambda) (\mathcal L^{global}_{local} +\mathcal L^{global}_{medium}), 
\label{eq:overall_CL}
\end{equation}
where $\lambda$ is a trade-off coefficient, the first three terms $\mathcal L^{local}_{ll'}$,  $\mathcal L^{medium}_{mm'}$ and $\mathcal L^{global}_{gg'}$ represent the peer-level semantic alignment objective functions, while the last two terms $\mathcal L^{global}_{local}$ and $\mathcal L^{global}_{medium}$ represent the cross-level semantic alignment objective functions.

\subsection{Softened Objective Function}
\label{sec:soften}
The one-hot label assumes there
is absolutely no similarity between different classes. However, in medical imaging, the images from different classes may have some local similarities (\eg the tissues and organs unrelated
to diseases). Thus, we propose the softened objective function to alleviate the negative effect caused by local similarities. We first define the corresponding softened label $\widetilde{\textbf{y}}_{i}$ as
\begin{equation}
  \widetilde{\textbf{y}}_{i} = (1-\alpha) {\textbf{y}}_{i} + \alpha/(N-1),
  \label{eq:dt}
\end{equation}
where $\textbf{y}_{i}$ is the original one-hot label, $\alpha$ is a smoothing hyper-parameter, and $N$ is the number of videos in a training batch. Then the softened cross-entropy (CE) loss $\mathcal L^{soften}$ with corresponding softened label $\widetilde{\textbf{y}}_{i}$ is formulated as 
\begin{equation}
\mathcal L^{soften} = \frac{1}{2N}\sum_{i=1}^{N}(CE(\textbf{o}_{i},\widetilde{\textbf{y}}_{i})+CE(\textbf{o}_{i}',\widetilde{\textbf{y}}_{i})),  
\label{eq:classification}
\end{equation}
where $\textbf{o}_{i} = f_{\theta}(\textbf{F}_{i}^{g})$, $\textbf{o}_{i}' = f_{\theta}(\textbf{F}_{i}^{g'})$, and $f_{\theta}$ is a linear classifier. 

Finally, the total loss can be written as 
\begin{equation}
\mathcal L = \mathcal L^{con} + \beta \mathcal L^{soften},
\label{eq:final}
\end{equation}
where the parameter $\beta$ is used to balance the total CL loss and softened CE loss. The whole algorithm is summarized in Algorithm~\ref{alg:algo}.

%===========================================================
\section{Experiments}
\label{sec:Experiment}

\subsection{Experimental Settings}

\subsubsection{Network Architective.}
In our experiments, we apply the widely used ResNet18-FPN~\cite{lin2017feature} network as the backbone. Conv2 to Conv5 are followed by a convolution layer with kernel size of 1*1 to obtain intermediate feature maps. In the FPN structure, we double upsampling the intermediate feature maps, and then add them to the intermediate feature maps of the previous layer. The intermediate feature maps are followed by a convolution layer with kernel size of 3*3 to obtain local embedding, medium embedding, and global embedding, respectively. All convolution layers are followed by batch normalization and ReLU. The projection heads $h$, $m$, and $g$ are all 1-layer MLP. After the projection heads, the local, medium, and global embeddings are reduced to 256-dimensional feature vectors for CL tasks. The linear classifier is a fully connected (FC) layer.

\subsubsection{Pretraining Details.} We use the US-4~\cite{chen2021uscl} video dataset (lung and liver) for pretraining and fine-tune the last 3 layers of pretrained models on various downstream tasks to evaluate the transferability of the proposed US video pretraining models. During the pretraining process, the input images are randomly cropped and resized to 224*224, followed by random flipping and color jitter. The pretraining epoch and batch size are set to 300 and 32, respectively. The parameters of models are obtained by optimizing the loss functions via an Adam optimizer with a learning rate $3 \times 10^{-4}$ and a weight decay rate $10^{-4}$. Following the popular CL evaluations~\cite{chen2020simple}, the backbone is used for fine-tuning on downstream tasks, projection heads ($h$, $m$, and $g$) and linear classifier ($f_{\theta}$) are discarded when the pretraining is completed. The $\tau\!=\!0.5$ is a tuning temperature parameter as in SimCLR~\cite{chen2020simple}. We empirically set $\lambda\!=\!0.5$, indicating that peer-level semantic alignment and cross-level semantic alignment have equal weights in the CL loss. The smoothing parameter $\alpha$ set to 0.2 indicates slight label smoothing. The $\beta$ is empirically set to 0.2, indicating that the CL loss dominates the total loss. 
All experiments were implemented using PyTorch and a single RTX 3090 GPU.

\subsubsection{Downstream Datasets.} We fine-tune our pretrained backbones on four US datasets (POCUS~\cite{born2021accelerating}, Thyroid US~\cite{pedraza2015open}, and BUSI-BUI~\cite{Walid2019DB,Rodrigues2018Mendeley} joint dataset), and a chest X-ray dataset (COVID-Xray-5k~\cite{minaee2020deep}) to evaluate the transferability of our pretraining models. For fair comparisons, all fine-tuning results on downstream datasets are obtained with 5-fold cross-validation. The POCUS is a lung convex probe US dataset for pneumonia detection that contains 2116 frames across 140 videos from three categories (COVID-19, bacterial pneumonia and the regular). The BUSI contains 780 breast tumor US images from three classes (the normal, benign and malignant), while BUI consists of 250 breast cancer US images with 100 benign and 150 malignant. Thyroid US~\cite{pedraza2015open} dataset contains thyroid images with 61 benign and 288 malignant. Note that the BUSI, BUI and Thyroid US datasets are collected with linear probes. The COVID-Xray-5k is a chest X-ray dataset that contains 2084 training and 3100 test images from two classes (COVID-19 and the normal). In our fine-tuning experiments, the learning rate, weight decay rate and epoch are set to $10^{-2}$, $10^{-4}$ and 30, respectively. The performance is assessed with Precision, Recall, Accuracy or F1 score.

\subsection{Ablation Studies}
\label{sec:ablation}
In this section, we verify the effectiveness of each component in our approach on the downstream POCUS pneumonia detection task, and all the models are pretrained on the US-4 dataset. 
\subsubsection{Peer-level Semantic Alignment.}

The impact of peer-level semantic alignment is summarized in Table~\ref{table:Ablation_Study_peer_level}. We reimplement the self-supervised method SimCLR~\cite{chen2020simple} with ResNet18 (w/o FPN) as our baseline. We can find that the baseline cannot achieve satisfying performance ($85.6\%$, $87.0\%$ and $86.9\%$ in terms of Precision, Recall and Accuracy, respectively). The effectiveness of fine-grained contrast, medium-grained contrast and coarse-grained contrast can be verified by comparing backbones pretrained using $\mathcal L_{ll'}^{local}$, $\mathcal L_{mm'}^{medium}$, $\mathcal L_{gg'}^{global}$ with the baseline, which contribute to the absolute performance gains
of $1.4\%$, $2.7\%$ and $3.2\%$ in terms of Accuracy. In addition, better or comparable results can be achieved when using two contrasts (\eg coarse-grained contrast and medium-grained contrast) than a single contrast (\eg coarse-grained contrast). Considering the excellent performance of backbones when using only the global CL loss (\ie an Accuracy of $90.1\%$), and using both the global CL loss and the medium CL loss (\ie an Accuracy of $91.9\%$), we argue that coarse-grained global information is very important to improve the transferability of pretrained US models. The best performance is achieved when all three peer-level semantic contrasts are used.

\setlength{\tabcolsep}{2pt}
\begin{table}[t]
\scriptsize
\begin{center}
\caption{
Impact of peer-level semantic alignment. The models are pretrained on US-4 and fine-tuned on POCUS dataset.
}
\label{table:Ablation_Study_peer_level}
\begin{tabular}{lccccc}
\toprule\noalign{\smallskip}
$\mathcal L_{ll'}^{local}$  & $\mathcal L_{mm'}^{medium}$ & $\mathcal L_{gg'}^{global}$ & Precision ($\%$) & Recall ($\%$) & Accuracy ($\%$)\\
\noalign{\smallskip}
\hline
\noalign{\smallskip}
 &  &  & 85.6  & 87.0  & 86.9\\
 
\checkmark &  &  & 87.6 & 87.5  & 88.3\\ 

& \checkmark &  & 87.9 & 90.2  & 89.6 \\

&   & \checkmark &  90.5  & 89.8  & 90.1 \\

\checkmark & \checkmark &  &  89.3 & 89.5  & 90.1 \\

\checkmark & & \checkmark & 91.0 & 90.7  & 90.5 \\

& \checkmark &  \checkmark &  91.9 & 92.3  & 91.9\\

\checkmark & \checkmark &  \checkmark &  91.3 & 92.3  & 92.0 \\
\bottomrule
\end{tabular}
\end{center}
\end{table}
\setlength{\tabcolsep}{0.1pt}

\setlength{\tabcolsep}{3pt}
\begin{table}[t]
\scriptsize
\begin{center}
\caption{
Impact of cross-level semantic alignment. The models are pretrained on US-4 and fine-tuned on the POCUS dataset.}
\label{table:Ablation_Study_cross_level}
\begin{tabular}{lcccc}
\toprule\noalign{\smallskip}
$\mathcal L^{global}_{medium}$ & $\mathcal L^{global}_{local}$ & Precision ($\%$) & Recall ($\%$) & Accuracy ($\%$)\\
\noalign{\smallskip}
\hline
\noalign{\smallskip}
 &  & 85.6  & 87.0  & 86.9\\
 
\checkmark &  &   88.0  & 89.6  & 89.4 \\

& \checkmark  &   90.4  & 91.1  & 90.8 \\

\checkmark & \checkmark  &   89.4  & 90.8  & 90.9\\
\bottomrule
\end{tabular}
\end{center}
\end{table}
\setlength{\tabcolsep}{0.1pt}

\subsubsection{Cross-level Semantic Alignment.}

The impact of cross-level semantic alignment is summarized in Table~\ref{table:Ablation_Study_cross_level}. In previous experiments, we find that global information is pivotal to CL tasks. Therefore, when performing cross-level semantic alignment, we regard the global feature as an \emph{anchor}, and only consider aligning the local features and the middle-level features with the global features (\ie global-local contrast and global-medium contrast). From Table~\ref{table:Ablation_Study_cross_level}, we can observe that consistent performance gains are achieved by conducting global-local contrast and global-medium contrast in terms of Precision, Recall and Accuracy. When global-local contrast and global-medium contrast are performed at the same time, our pretrained model can achieve the best performance in terms of Accuracy (\ie $90.9\%$). However, using the proposed peer-level semantic alignment can achieve the best Accuracy of $92.0\%$ as shown in Table~\ref{table:Ablation_Study_peer_level}. Therefore, it is difficult to learn better transferable representations by using cross-level semantic alignment alone. Next, we will demonstrate that using the peer-level semantic alignment and cross-level semantic alignment at the same time can bring significant performance gains.

\setlength{\tabcolsep}{1.0pt}
\begin{table}[t]
\scriptsize
\begin{center}
\caption{
Ablation study of each component in our approach. The models are pretrained on US-4 and fine-tuned on POCUS dataset.}
\label{table:Ablation_Study}
\begin{tabular}{lcccccccc}
\toprule\noalign{\smallskip}
$\mathcal L_{ll'}^{local}$  & $\mathcal L_{mm'}^{medium}$ & $\mathcal L_{gg'}^{global}$ & $\mathcal L^{global}_{medium}$ & $\mathcal L^{global}_{local}$ & $\mathcal L^{soften}$ & Precision ($\%$) & Recall ($\%$) & Accuracy ($\%$)\\
\noalign{\smallskip}
\hline
\noalign{\smallskip}
 &  &  &  &  &  & 85.6  & 87.0  & 86.9\\
\checkmark &  &  &  &  &  & 87.6 & 87.5  & 88.3 \\
\checkmark & \checkmark &  &  &  &  &  89.3 & 89.5  & 90.1 \\
\checkmark & \checkmark & \checkmark &  &  &  &  91.3 & 92.3  & 92.0  \\
%\hline
\checkmark & \checkmark & \checkmark  & \checkmark &  &  &  92.2 & 93.2  & 93.2\\
\checkmark & \checkmark & \checkmark  &  \checkmark & \checkmark &  &  93.5  &  94.2  & 94.4\\
\checkmark & \checkmark & \checkmark  &  \checkmark & \checkmark &  \checkmark &  94.6 & 94.9  & 94.7\\
\bottomrule
\end{tabular}
\end{center}
\end{table}
\setlength{\tabcolsep}{0.1pt}

\begin{figure}[t]
\centering
\includegraphics[width=110mm]{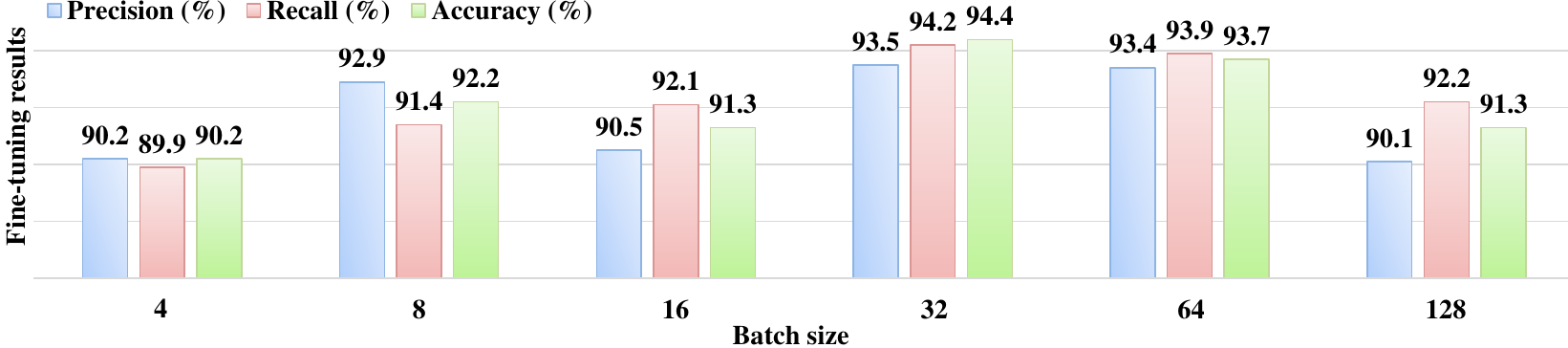} 
\caption{Impact of batch size. Fine-tuning results obtained on POCUS dataset with different batch sizes.}
\label{fig:batchsize}
\end{figure} 

\subsubsection{Ablation Study of Each Component.}
The impact of each component in our approach is summarized in Table~\ref{table:Ablation_Study}. We can see that each component can bring a certain performance improvement to our final method, which verifies the effectiveness of each component. An interesting observation is that our method (an Accuracy of $94.4\%$) without using labels can surpass the current state-of-the-art semi-supervised method USCL (an Accuracy of $94.2\%$) and supervised method (an Accuracy of $85.0\%$) on POCUS dataset (see Tables~\ref{table:Ablation_Study} and~\ref{table:POCUS}). This result demonstrates the superiority of peer-level semantic alignment and cross-level semantic alignment for learning transferable representations. In addition, the softened objective function can further facilitate the performance of our approach (from $94.4\%$ to $94.7\%$ in terms of Accuracy).

\subsubsection{Impact of Batch Size.} The effect of batch size is shown in Fig.~\ref{fig:batchsize}. We can observe that as the batch size increases, the overall performance demonstrates a trend from rising to decline, where the best overall performance is achieved when the batch size is 32. Compared with existing state-of-the-art CL algorithms that rely on large batch sizes (more negative samples), \eg 1024 in MoCo ~\cite{he2020momentum}, 2048 in SimCLR~\cite{chen2020simple}, to achieve convergence, our proposed method is easier to optimize. In this way, we can train the model in fewer epochs and steps to obtain a given accuracy. We argue that the multi-level contrast promotes the effective and sufficient interaction of information from different layers of a DNN, thus we can use a smaller batch size.
 
\subsubsection{Impact of Label Rate.} 
\begin{figure}[t]
\centering
\includegraphics[width=78mm]{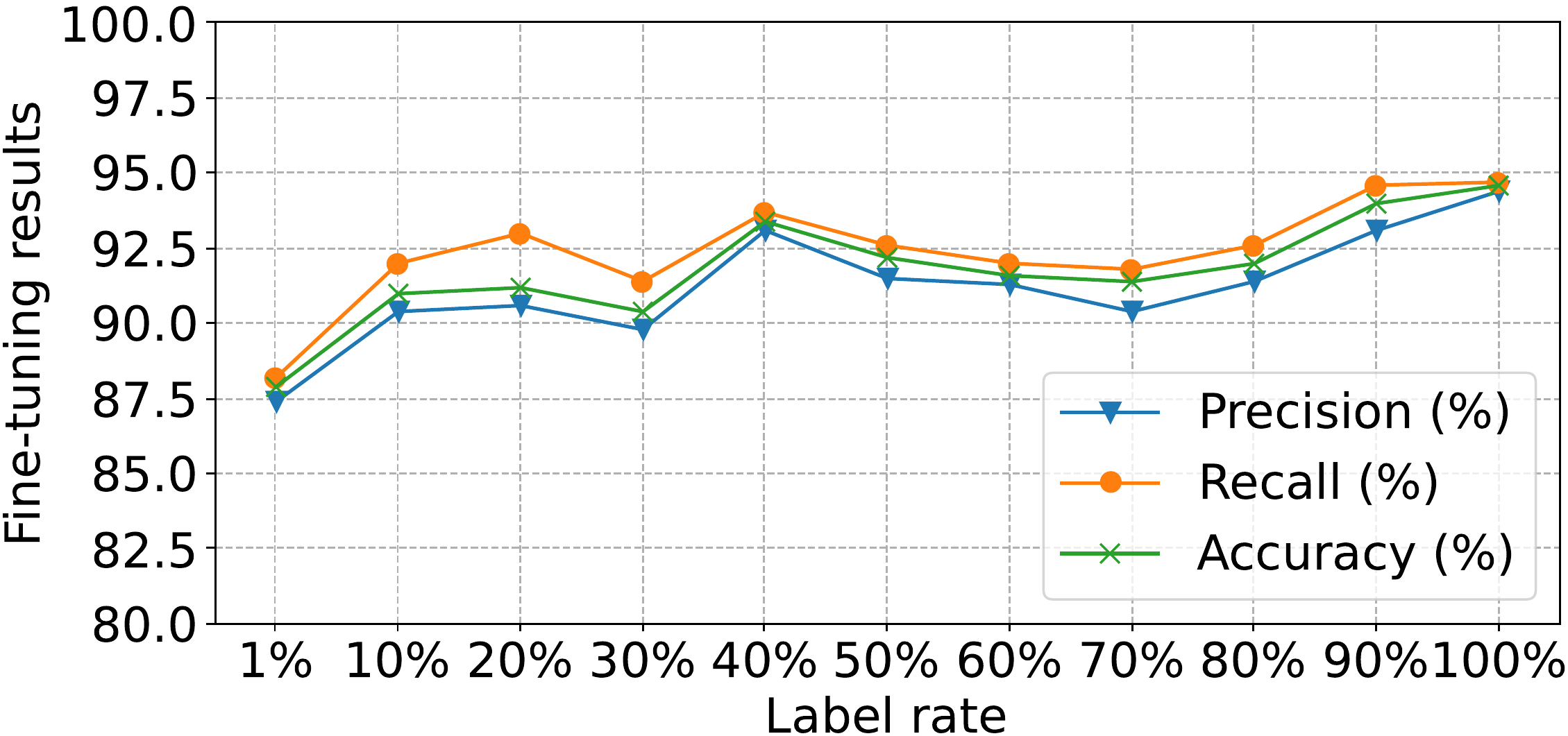} 
\caption{Impact of the label rate. Fine-tuning results obtained on POCUS dataset with different label rates.}
\label{fig:labelrate}
\end{figure}

The effect of label rate is shown in Fig.~\ref{fig:labelrate}. We pretrain eleven models on US-4 dataset with different label rates. We find that the performance of models gradually improve with the increase of label, as more label can bring stronger supervision signals to promote the process of pretraining.

\subsection{Comparison with State-of-the-art Methods}
To verify the effectiveness of our approach, we compare the proposed HiCo with supervised ResNet18 backbones (\ie ``ImageNet'' pretrained on  ImageNet dataset, ``Supervised'' pretrained on US-4 dataset), and other backbones pretrained on US-4 dataset with semi-supervised methods (\ie Temporal Ensembling (TE)~\cite{laine2016temporal}, $\Pi$  Model~\cite{laine2016temporal}, FixMatch~\cite{sohn2020fixmatch}, USCL~\cite{chen2021uscl}) and self-supervised methods (\ie MoCo v2~\cite{chen2020improved}, SimCLR~\cite{chen2020simple}). 

\subsubsection{POCUS Pneumonia Detection.} On POCUS, we fine-tune the last three layers to verify the transferability of pretrained backbones on the pneumonia detection task. The results are summarized in Table~\ref{table:POCUS}. We report the Accuracy of three classes (\ie COVID-19, pneumonia and the regular), total Accuracy and F1 scores on POCUS. The proposed HiCo achieves the best performance in terms of total Accuracy (\ie $94.7\%$) and F1 ($94.6\%$) scores, which are significantly better than other supervised, semi-supervised and self-supervised methods. HiCo also obtains the best Accuracy on COVID-19 and pneumonia classes, and the second-best Accuracy on the regular. USCL achieves the best Accuracy on the regular, which is a semi-supervised CL method with a carefully designed sample pair generation strategy for US video sampling. Compared with USCL, HiCo makes full use of the peer-level and cross-level knowledge from the network via multi-level contrast, which presents a stronger representation capability.

\setlength{\tabcolsep}{1pt}
\begin{table}[t]
\scriptsize
\begin{center}
\caption{
Comparison of fine-tuning results on POCUS and BUSI-BUI classification datasets. Top two results are in bold and underlined.}
\label{table:POCUS}
\begin{tabular}{lcccccc}
\toprule\noalign{\smallskip}
\multirow{2}{*}{Method} & \multicolumn{5}{c}{{POCUS}} &
			 \multicolumn{1}{c}{BUSI-BUI}\\
             \cmidrule(r){2-6} \cmidrule(r){7-7}
				\multicolumn{1}{c}{} & COVID-19 & Pneumonia & Regular & Accuracy ($\%$) & F1 ($\%$) &  Accuracy ($\%$) \\
\noalign{\smallskip}
\hline
\noalign{\smallskip}
ImageNet~\cite{russakovsky2015imagenet} & 79.5 & 78.6 & 88.6 & 84.2 & 81.8 & 84.9\\

Supervised~\cite{chen2021uscl} & 83.7 & 82.1 & 86.5 & 85.0 & 82.8 & 71.3\\
\hline

TE~\cite{laine2016temporal} & 75.7 & 70.0 & 89.4 & 81.7 & 79.0 & 71.8\\

$\Pi$  Model~\cite{laine2016temporal} & 77.6 & 76.4 & 88.7 & 83.2 & 80.6 & 69.7\\

FixMatch~\cite{sohn2020fixmatch} & 83.0 & 77.5 & 85.7 & 83.6 & 81.6 & 70.3\\

MoCo v2~\cite{chen2020improved} & 79.7 & 81.4 & 88.9 & 84.8 & 82.8 & 77.8\\

SimCLR~\cite{chen2020simple} & 83.2 & 89.4 & 87.1 & 86.4 & 86.3 & 74.6\\

USCL~\cite{chen2021uscl} & \underline{90.8} & \underline{97.0} & \textbf{95.4} & \underline{94.2} & \underline{94.0} & \underline{85.5}\\

\textbf{HiCo (Ours)} & \textbf{97.1} & \textbf{100.0} & \underline{92.5} & \textbf{94.7} & \textbf{94.6} & \textbf{86.0}\\
%\hline
\bottomrule
\end{tabular}
\end{center}
\end{table}
\setlength{\tabcolsep}{0.1pt}

\subsubsection{BUSI-BUI Breast Cancer Classification.} The fine-tuning results on BUSI-BUI joint dataset are summarized in Table~\ref{table:POCUS}. Among the compared methods, HiCo provides the best Accuracy (\ie $86.0\%$). Compared with ``Supervised'', HiCo obtains an absolute gain of $14.7\%$ in terms of Accuracy. We also observe that our HiCo does not demonstrate significant superiority to ``ImageNet'' like on POCUS dataset. This is because our pretraining dataset US-4 is captured with convex probes, while BUSI-BUI joint dataset is captured with linear probes. The domain gap between convex and linear probes damages the performance of backbones pretrained on convex probe US data including our HiCo.

\subsection{Transferability to Thyroid US Images}
We further evaluate the transferability of HiCo on Thyroid US classification dataset. We compare HiCo with the other three backbones pretrained using supervised learning (\ie Supervised), learning from scratch (\ie LFS), and vanilla contrastive learning (\ie VCL). For fair comparisons, all the backbones are pretrained on US-4 dataset and fine-tuned the last three layers. The results are shown in Table~\ref{table:thyroid}. We report the Precision and Recall of two classes (\ie the benign and malignant) and the total Accuracy. We find that HiCo has the consistent best performance on the classification of two classes, and its total Accuracy of $90.5\%$ is also significantly better than the other three methods.

\setlength{\tabcolsep}{6pt}
\begin{table}[t]
\scriptsize
\begin{center}
\caption{
Comparison of fine-tuning results on Thyroid US Images dataset. ``Supervised'', ``LFS'', and ``VCL'' denote the backbones pretrained with supervised learning, learning from scratch, and vanilla contrastive learning, respectively.}
\label{table:thyroid}
\begin{tabular}{lcccccc|}
\toprule
\multirow{2}{*}{Method} & \multicolumn{2}{c}{{Precision ($\%$)}} &
			 \multicolumn{2}{c}{Recall ($\%$)} & \multirow{2}{*}{Accuracy ($\%$)} \\
             \cmidrule(r){2-3} \cmidrule(r){4-5}
				\multicolumn{1}{c}{} & Benign & Malignant & Benign & Malignant & \\
\noalign{\smallskip}
\hline
\noalign{\smallskip}
Supervised & 81.3 & 89.0 & 42.6 & 97.9 & 88.3  \\

LFS & 79.3 & 88.1 & 37.7 & 97.9 & 87.4  \\

VCL & 89.3 & 88.8 & 41.0 & 99.0 & 88.8  \\

\textbf{HiCo (Ours)} & \textbf{91.2} & \textbf{89.7} & \textbf{49.2} & \textbf{99.7} & \textbf{90.5} \\
\bottomrule
\end{tabular}
\end{center}
\end{table}
\setlength{\tabcolsep}{1.4pt}

\subsection{Cross-modal Transferability to Chest X-ray}

The goal of this work is to design an effective method for US video model pretraining. To test our approach's transferability, we also apply our approach to a Chest X-ray classification dataset (\ie COVID-Xray-5k). This experiment can verify the cross-modal transferability of our approach. The detailed results about Supervised, LFS and VCL are listed in Table~\ref{table:COVID-Xray-5k}. From Table~\ref{table:COVID-Xray-5k}, we can see that our HiCo achieves the best performance of two classes (\ie COVID-19 and the normal) and total Accuracy. Specifically, HiCo outperforms the LFS, VCL and Supervised by $5.0\%$, $2.6\%$ and $2.1\%$, respectively. Although our approach is designed for US video model pretraining, the above results demonstrate its excellent cross-modal transferability.

\setlength{\tabcolsep}{6pt}
\begin{table}[t]
\scriptsize
\begin{center}
\caption{
Comparison of fine-tuning results on COVID-Xray-5k chest X-ray dataset. ``Supervised'', ``LFS'', and ``VCL'' denote the backbones pretrained with supervised learning, learning from scratch, and vanilla contrastive learning, respectively.}
\label{table:COVID-Xray-5k}
\begin{tabular}{lcccccc}
\toprule
\multirow{2}{*}{Method} & \multicolumn{2}{c}{{Precision ($\%$)}} &
\multicolumn{2}{c}{Recall ($\%$)} & \multirow{2}{*}{Accuracy ($\%$)} \\
\cmidrule(r){2-3} \cmidrule(r){4-5}
\multicolumn{1}{c}{} & COVID-19 & Normal & COVID-19 & Normal & \\
\noalign{\smallskip}
\hline
\noalign{\smallskip}
Supervised & 94.2 & 94.5 & 90.6 & 96.7 & 94.4\\

LFS & 87.2 & 94.2 & 90.5 & 92.1 & 91.5\\

VCL & 92.2 & 94.9 & 91.5 & 95.4 & 93.9 \\

\textbf{HiCo (Ours)} & \textbf{94.5} & \textbf{97.7} & \textbf{96.2} & \textbf{96.7} & \textbf{96.5} \\
\bottomrule
\end{tabular}
\end{center}
\end{table}
\setlength{\tabcolsep}{1.4pt}

\section{Conclusion}
In this work, we propose the hierarchical contrastive learning for US video model pretraining, which fully and efficiently utilizes both peer-level and cross-level knowledge from a DNN via multi-level contrast, leading to the remarkable transferability for the pretrained model. The advantage of our proposed method is that it flexibly extends the existing CL architecture (\ie the vanilla contrastive learning framework) and promotes knowledge communication inside the network by designing several simple and effective loss functions instead of designing a complex network structure. We empirically identify that multi-level contrast can greatly accelerate the convergence speed of pretrained models in downstream tasks, and improve the representation ability of models. In addition, a softened objective function is introduced to alleviate the negative effect of some local similarities between different classes, which further facilitates the CL process. Future works include exploiting more general frameworks for multi-level contrast and other applications for US diagnosis.

\myPara{Acknowledgments.} This work is supported by the National Natural Science Foundation of China (No. 62101351), the Guangdong Basic and Applied Basic Research Foundation (No.2020A1515110376), Shenzhen Outstanding Scientific and Technological Innovation Talents PhD Startup Project (No. RCBS202106091\\04447108), the National Key R $\&$ D Program of China under Grant (No.2021ZD01\\13400), Guangdong Provincial Key Laboratory of Big Data Computing, and the Chinese University of Hong Kong, Shenzhen.

%===========================================================
\bibliographystyle{splncs}
\bibliography{HiCo}

\begin{thebibliography}{10}

\bibitem{born2021accelerating}
Born, J., Wiedemann, N., Cossio, M., Buhre, C., Br{\"a}ndle, G., Leidermann,
  K., Goulet, J., Aujayeb, A., Moor, M., Rieck, B.,  et~al.:
\newblock Accelerating detection of lung pathologies with explainable
  ultrasound image analysis.
\newblock Applied Sciences \textbf{11} (2021)  672

\bibitem{gao2016describing}
Gao, Y., Maraci, M.A., Noble, J.A.:
\newblock Describing ultrasound video content using deep convolutional neural
  networks.
\newblock In: 2016 IEEE 13th International Symposium on Biomedical Imaging.
  (2016)  787--790

\bibitem{chen2021uscl}
Chen, Y., Zhang, C., Liu, L., Feng, C., Dong, C., Luo, Y., Wan, X.:
\newblock Uscl: Pretraining deep ultrasound image diagnosis model through video
  contrastive representation learning.
\newblock In: International Conference on Medical Image Computing and
  Computer-Assisted Intervention, Springer (2021)  627--637

\bibitem{liu2020semi}
Liu, L., Lei, W., Wan, X., Liu, L., Luo, Y., Feng, C.:
\newblock Semi-supervised active learning for covid-19 lung ultrasound
  multi-symptom classification.
\newblock In: 2020 IEEE 32nd International Conference on Tools with Artificial
  Intelligence (ICTAI), IEEE (2020)  1268--1273

\bibitem{gao2021multi}
Gao, L., Zhou, R., Dong, C., Feng, C., Li, Z., Wan, X., Liu, L.:
\newblock Multi-modal active learning for automatic liver fibrosis diagnosis
  based on ultrasound shear wave elastography.
\newblock In: 2021 IEEE 18th International Symposium on Biomedical Imaging
  (ISBI), IEEE (2021)  410--414

\bibitem{su2020convolutional}
Su, H., Chang, Z., Yu, M., Gao, J., Li, X., Zheng, S.,  et~al.:
\newblock Convolutional neural network with adaptive inferential framework for
  skeleton-based action recognition.
\newblock Journal of Visual Communication and Image Representation \textbf{73}
  (2020)  102925

\bibitem{jiao2020self}
Jiao, J., Droste, R., Drukker, L., Papageorghiou, A.T., Noble, J.A.:
\newblock Self-supervised representation learning for ultrasound video.
\newblock In: 2020 IEEE 17th International Symposium on Biomedical Imaging.
  (2020)  1847--1850

\bibitem{zhang2016colorful}
Zhang, R., Isola, P., Efros, A.A.:
\newblock Colorful image colorization.
\newblock In: European conference on computer vision, Springer (2016)  649--666

\bibitem{he2020momentum}
He, K., Fan, H., Wu, Y., Xie, S., Girshick, R.:
\newblock Momentum contrast for unsupervised visual representation learning.
\newblock In: Proceedings of the IEEE/CVF conference on computer vision and
  pattern recognition. (2020)  9729--9738

\bibitem{chen2020simple}
Chen, T., Kornblith, S., Norouzi, M., Hinton, G.:
\newblock A simple framework for contrastive learning of visual
  representations.
\newblock In: International conference on machine learning, PMLR (2020)
  1597--1607

\bibitem{jiao2020self2}
Jiao, J., Cai, Y., Alsharid, M., Drukker, L., Papageorghiou, A.T., Noble, J.A.:
\newblock Self-supervised contrastive video-speech representation learning for
  ultrasound.
\newblock In: International Conference on Medical Image Computing and
  Computer-Assisted Intervention, Springer (2020)  534--543

\bibitem{zhang2020contrastive}
Zhang, Y., Jiang, H., Miura, Y., Manning, C.D., Langlotz, C.P.:
\newblock Contrastive learning of medical visual representations from paired
  images and text.
\newblock arXiv preprint arXiv:2010.00747 (2020)

\bibitem{chen2020improved}
Chen, X., Fan, H., Girshick, R., He, K.:
\newblock Improved baselines with momentum contrastive learning.
\newblock arXiv preprint arXiv:2003.04297 (2020)

\bibitem{gao2020accurate}
Gao, J., Shi, Z., Wang, G., Li, J., Yuan, Y., Ge, S., Zhou, X.:
\newblock Accurate temporal action proposal generation with relation-aware
  pyramid network.
\newblock In: Proceedings of the AAAI Conference on Artificial Intelligence.
  Volume~34. (2020)  10810--10817

\bibitem{ren2015faster}
Ren, S., He, K., Girshick, R., Sun, J.:
\newblock Faster r-cnn: Towards real-time object detection with region proposal
  networks.
\newblock IEEE Trans on Pattern Analysis and Machine Intelligence \textbf{39}
  (2016)  1137--1149

\bibitem{xu2022seed}
Xu, H., Zhang, X., Li, H., Xie, L., Dai, W., Xiong, H., Tian, Q.:
\newblock Seed the views: Hierarchical semantic alignment for contrastive
  representation learning.
\newblock IEEE Transactions on Pattern Analysis and Machine Intelligence (2022)

\bibitem{lee2020contrastive}
Lee, S., Lee, D.B., Hwang, S.J.:
\newblock Contrastive learning with adversarial perturbations for conditional
  text generation.
\newblock 9th International Conference on Learning Representations, {ICLR}
  (2021)

\bibitem{li2022keywords}
Li, M., Lin, X., Chen, X., Chang, J., Zhang, Q., Wang, F., Wang, T., Liu, Z.,
  Chu, W., Zhao, D.,  et~al.:
\newblock Keywords and instances: A hierarchical contrastive learning framework
  unifying hybrid granularities for text generation.
\newblock (2022)  4432--4441

\bibitem{li2022hiclre}
Li, D., Zhang, T., Hu, N., Wang, C., He, X.:
\newblock Hiclre: A hierarchical contrastive learning framework for distantly
  supervised relation extraction.
\newblock (2022)  2567--2578

\bibitem{wang2022heloc}
Wang, X., Wu, Q., Zhang, H., Lyu, C., Jiang, X., Zheng, Z., Lyu, L., Hu, S.:
\newblock Heloc: Hierarchical contrastive learning of source code
  representation.
\newblock arXiv preprint arXiv:2203.14285 (2022)

\bibitem{schmarje2020survey}
Schmarje, L., Santarossa, M., Schr{\"o}der, S.M., Koch, R.:
\newblock A survey on semi-, self-and unsupervised techniques in image
  classification.
\newblock arXiv preprint arXiv:2002.08721 (2020)

\bibitem{he2016deep}
He, K., Zhang, X., Ren, S., Sun, J.:
\newblock Deep residual learning for image recognition.
\newblock In: CVPR. (2016)  770--778

\bibitem{gao2022efficient}
Gao, J., Sun, X., Ghanem, B., Zhou, X., Ge, S.:
\newblock Efficient video grounding with which-where reading comprehension.
\newblock IEEE Transactions on Circuits and Systems for Video Technology (2022)

\bibitem{chi2017thyroid}
Chi, J., Walia, E., Babyn, P., Wang, J., Groot, G., Eramian, M.:
\newblock Thyroid nodule classification in ultrasound images by fine-tuning
  deep convolutional neural network.
\newblock Journal of digital imaging \textbf{30} (2017)  477--486

\bibitem{yap2017automated}
Yap, M.H., Pons, G., Marti, J., Ganau, S., Sentis, M., Zwiggelaar, R., Davison,
  A.K., Marti, R.:
\newblock Automated breast ultrasound lesions detection using convolutional
  neural networks.
\newblock IEEE journal of biomedical and health informatics \textbf{22} (2017)
  1218--1226

\bibitem{huang2017breast}
Huang, Q., Luo, Y., Zhang, Q.:
\newblock Breast ultrasound image segmentation: a survey.
\newblock International journal of computer assisted radiology and surgery
  \textbf{12} (2017)  493--507

\bibitem{Walid2019DB}
Al-Dhabyani, W., Gomaa, M., Khaled, H., Fahmy, A.:
\newblock Dataset of breast ultrasound images.
\newblock Data in Brief \textbf{28} (2019)

\bibitem{Rodrigues2018Mendeley}
Rodrigues, P.S.:
\newblock Breast ultrasound image.
\newblock Mendeley Data, V1, doi: 10.17632/wmy84gzngw.1 (2018)

\bibitem{pedraza2015open}
Pedraza, L., Vargas, C., Narv{\'a}ez, F., Dur{\'a}n, O., Mu{\~n}oz, E., Romero,
  E.:
\newblock An open access thyroid ultrasound image database.
\newblock In: SPIE. Volume 9287. (2015)

\bibitem{nguyen2020ultrasound}
Nguyen, D.T., Kang, J.K., Pham, T.D., Batchuluun, G., Park, K.R.:
\newblock Ultrasound image-based diagnosis of malignant thyroid nodule using
  artificial intelligence.
\newblock Sensors \textbf{20} (2020)  1822

\bibitem{li2020automated}
Li, P., Zhao, H., Liu, P., Cao, F.:
\newblock Automated measurement network for accurate segmentation and parameter
  modification in fetal head ultrasound images.
\newblock Medical Biological Engineering Computing \textbf{58} (2020)
  2879--2892

\bibitem{kalafat2020lung}
Kalafat, E., Yaprak, E., Cinar, G., Varli, B., Ozisik, S., Uzun, C., Azap, A.,
  Koc, A.:
\newblock Lung ultrasound and computed tomographic findings in pregnant woman
  with covid-19.
\newblock Ultrasound in Obstetrics \& Gynecology \textbf{55} (2020)  835--837

\bibitem{long2017lung}
Long, L., Zhao, H.T., Zhang, Z.Y., Wang, G.Y., Zhao, H.L.:
\newblock Lung ultrasound for the diagnosis of pneumonia in adults: a
  meta-analysis.
\newblock Medicine \textbf{96} (2017)

\bibitem{kerdegari2021automatic}
Kerdegari, H., Nhat, P.T.H., McBride, A., Razavi, R., Van~Hao, N., Thwaites,
  L., Yacoub, S., Gomez, A.:
\newblock Automatic detection of b-lines in lung ultrasound videos from severe
  dengue patients.
\newblock In: 2021 IEEE 18th International Symposium on Biomedical Imaging.
  (2021)  989--993

\bibitem{wang2019quantifying}
Wang, X., Burzynski, J.S., Hamilton, J., Rao, P.S., Weitzel, W.F., Bull, J.L.:
\newblock Quantifying lung ultrasound comets with a convolutional neural
  network: Initial clinical results.
\newblock Computers in biology and medicine \textbf{107} (2019)  39--46

\bibitem{carrer2020automatic}
Carrer, L., Donini, E., Marinelli, D., Zanetti, M., Mento, F., Torri, E.,
  Smargiassi, A., Inchingolo, R., Soldati, G., Demi, L.,  et~al.:
\newblock Automatic pleural line extraction and covid-19 scoring from lung
  ultrasound data.
\newblock IEEE Transactions on Ultrasonics, Ferroelectrics, and Frequency
  Control \textbf{67} (2020)  2207--2217

\bibitem{xu2020boundary}
Xu, Y., Zhang, Y., Bi, K., Ning, Z., Xu, L., Shen, M., Deng, G., Wang, Y.:
\newblock Boundary restored network for subpleural pulmonary lesion
  segmentation on ultrasound images at local and global scales.
\newblock Journal of Digital Imaging \textbf{33} (2020)  1155--1166

\bibitem{6823625}
Razaak, M., Martini, M.G., Savino, K.:
\newblock A study on quality assessment for medical ultrasound video compressed
  via hevc.
\newblock IEEE Journal of Biomedical and Health Informatics \textbf{18} (2014)
  1552--1559

\bibitem{kwitt2013localizing}
Kwitt, R., Vasconcelos, N., Razzaque, S., Aylward, S.:
\newblock Localizing target structures in ultrasound video--a phantom study.
\newblock Medical image analysis \textbf{17} (2013)  712--722

\bibitem{ronneberger2015u}
Ronneberger, O., Fischer, P., Brox, T.:
\newblock U-net: Convolutional networks for biomedical image segmentation.
\newblock In: International Conference on Medical image computing and
  computer-assisted intervention, Springer (2015)  234--241

\bibitem{chen2014semantic}
Chen, L.C., Papandreou, G., Kokkinos, I., Murphy, K., Yuille, A.L.:
\newblock Semantic image segmentation with deep convolutional nets and fully
  connected crfs.
\newblock arXiv preprint arXiv:1412.7062 (2014)

\bibitem{chen2018encoder}
Chen, L.C., Zhu, Y., Papandreou, G., Schroff, F., Adam, H.:
\newblock Encoder-decoder with atrous separable convolution for semantic image
  segmentation.
\newblock In: European conference on computer vision. (2018)  801--818

\bibitem{cao2021swin}
Cao, H., Wang, Y., Chen, J., Jiang, D., Zhang, X., Tian, Q., Wang, M.:
\newblock Swin-unet: Unet-like pure transformer for medical image segmentation.
\newblock arXiv preprint arXiv:2105.05537 (2021)

\bibitem{weng2019unet}
Weng, Y., Zhou, T., Li, Y., Qiu, X.:
\newblock Nas-unet: Neural architecture search for medical image segmentation.
\newblock IEEE Access \textbf{7} (2019)  44247--44257

\bibitem{huang2022extracting}
Huang, R., Ying, Q., Lin, Z., Zheng, Z., Tan, L., Tang, G., Zhang, Q., Luo, M.,
  Yi, X., Liu, P.,  et~al.:
\newblock Extracting keyframes of breast ultrasound video using deep
  reinforcement learning.
\newblock Medical Image Analysis (2022)  102490

\bibitem{gong2021diagnosis}
Gong, B., Shi, J., Han, X., Zhang, H., Huang, Y., Hu, L., Wang, J., Du, J.,
  Shi, J.:
\newblock Diagnosis of infantile hip dysplasia with b-mode ultrasound via
  two-stage meta-learning based deep exclusivity regularized machine.
\newblock IEEE Journal of Biomedical and Health Informatics \textbf{26} (2021)
  334--344

\bibitem{korbar2018cooperative}
Korbar, B., Tran, D., Torresani, L.:
\newblock Cooperative learning of audio and video models from self-supervised
  synchronization.
\newblock Advances in Neural Information Processing Systems \textbf{31} (2018)

\bibitem{oord2018representation}
Oord, A.v.d., Li, Y., Vinyals, O.:
\newblock Representation learning with contrastive predictive coding.
\newblock arXiv (2018)

\bibitem{ye2019unsupervised}
Ye, M., Zhang, X., Yuen, P.C., Chang, S.F.:
\newblock Unsupervised embedding learning via invariant and spreading instance
  feature.
\newblock In: Proceedings of the IEEE/CVF Conference on Computer Vision and
  Pattern Recognition. (2019)  6210--6219

\bibitem{noroozi2016unsupervised}
Noroozi, M., Favaro, P.:
\newblock Unsupervised learning of visual representations by solving jigsaw
  puzzles.
\newblock In: European conference on computer vision, Springer (2016)  69--84

\bibitem{chen2019self}
Chen, L., Bentley, P., Mori, K., Misawa, K., Fujiwara, M., Rueckert, D.:
\newblock Self-supervised learning for medical image analysis using image
  context restoration.
\newblock Medical image analysis \textbf{58} (2019)  101539

\bibitem{doersch2015unsupervised}
Doersch, C., Gupta, A., Efros, A.A.:
\newblock Unsupervised visual representation learning by context prediction.
\newblock In: Proceedings of the IEEE international conference on computer
  vision. (2015)  1422--1430

\bibitem{gutmann2010noise}
Gutmann, M., Hyv{\"a}rinen, A.:
\newblock Noise-contrastive estimation: A new estimation principle for
  unnormalized statistical models.
\newblock In: Proceedings of the thirteenth international conference on
  artificial intelligence and statistics, JMLR Workshop and Conference
  Proceedings (2010)  297--304

\bibitem{grill2020bootstrap}
Grill, J.B., Strub, F., Altch{\'e}, F., Tallec, C., Richemond, P., Buchatskaya,
  E., Doersch, C., Avila~Pires, B., Guo, Z., Gheshlaghi~Azar, M.,  et~al.:
\newblock Bootstrap your own latent-a new approach to self-supervised learning.
\newblock Advances in neural information processing systems \textbf{33} (2020)
  21271--21284

\bibitem{chen2021exploring}
Chen, X., He, K.:
\newblock Exploring simple siamese representation learning.
\newblock In: Proceedings of the IEEE/CVF Conference on Computer Vision and
  Pattern Recognition. (2021)  15750--15758

\bibitem{chen2021empirical}
Chen, X., Xie, S., He, K.:
\newblock An empirical study of training self-supervised vision transformers.
\newblock In: Proceedings of the IEEE/CVF International Conference on Computer
  Vision. (2021)  9640--9649

\bibitem{lin2017feature}
Lin, T.Y., Doll{\'a}r, P., Girshick, R., He, K., Hariharan, B., Belongie, S.:
\newblock Feature pyramid networks for object detection.
\newblock In: Proceedings of the IEEE conference on computer vision and pattern
  recognition. (2017)  2117--2125

\bibitem{minaee2020deep}
Minaee, S., Kafieh, R., Sonka, M., Yazdani, S., Soufi, G.J.:
\newblock Deep-covid: Predicting covid-19 from chest x-ray images using deep
  transfer learning.
\newblock Medical image analysis \textbf{65} (2020)  101794

\bibitem{laine2016temporal}
Laine, S., Aila, T.:
\newblock Temporal ensembling for semi-supervised learning.
\newblock In: ICLR. (2017)

\bibitem{sohn2020fixmatch}
Sohn, K., Berthelot, D., Li, C., Zhang, Z., Carlini, N., Cubuk, E.D., Kurakin,
  A., Zhang, H., Raffel, C.:
\newblock Fixmatch: Simplifying semi-supervised learning with consistency and
  confidence.
\newblock arXiv:2001.07685 (2020)

\bibitem{russakovsky2015imagenet}
Russakovsky, O., Deng, J., Su, H., Krause, J., Satheesh, S., Ma, S., Huang, Z.,
  Karpathy, A., Khosla, A., Bernstein, M.,  et~al.:
\newblock Imagenet large scale visual recognition challenge.
\newblock International Journal of Computer Vision \textbf{115} (2015)
  211--252

\end{thebibliography}

\end{document}